\documentclass{article}

\usepackage{amsmath}
\usepackage{amssymb}
\usepackage{booktabs}
\usepackage{pifont}
\usepackage{caption, subcaption}
\usepackage{float}
\usepackage{algorithm}
\usepackage{algorithmic}
\usepackage{graphicx}
\usepackage{multirow}
\usepackage{appendix}
\usepackage{etoolbox}
\usepackage{arydshln}
\usepackage{xcolor}

\DeclareMathOperator*{\argmin}{arg\,min}

\usepackage[final]{corl_2020} 
 
\newcommand{\cmark}{\ding{51}}%
\newcommand{\xmark}{\ding{55}}%

\title{Map-Adaptive Goal-Based Trajectory Prediction}

\author{
  Lingyao Zhang, Po-Hsun Su, Jerrick Hoang, Galen Clark Haynes, Micol Marchetti-Bowick\\[4pt]
  Uber Advanced Technologies Group, Pittsburgh, PA\\[3pt]
  \texttt{\{lingyaoz,evan.su,jhoang,gch,mmarchettibowick\}@uber.com}\\
}

\begin{document}

\maketitle


\begin{abstract}
We present a new method for multi-modal, long-term vehicle trajectory prediction. Our approach relies on using lane centerlines captured in rich maps of the environment to generate a set of proposed goal paths for each vehicle. Using these paths -- which are generated at run time and therefore dynamically adapt to the scene -- as spatial anchors, we predict a set of goal-based trajectories along with a categorical distribution over the goals. This approach allows us to directly model the goal-directed behavior of traffic actors, which unlocks the potential for more accurate long-term prediction. 
Our experimental results on both a large-scale internal driving dataset and on the public nuScenes dataset show that our model outperforms state-of-the-art approaches for vehicle trajectory prediction over a \mbox{6-second horizon}. We also empirically demonstrate that our model is better able to generalize to road scenes from a completely new city than existing methods. 
\end{abstract}


\section{Introduction}
\vspace{-.1cm}

Predicting the future behavior of actors in a traffic scene is a critical task for the development of safe and effective self-driving technology. In this work, we focus on predicting the future motion of vehicles using rich map context. In particular, we leverage the fact that human drivers exhibit \emph{goal-directed behavior}, meaning they drive with the aim of reaching a particular destination. Furthermore, we recognize that the motion of drivers is heavily guided by the network of roads and lanes, which we access via high-definition maps of the environment. Using these observations, we construct a multi-modal trajectory prediction model that first generates a set of \emph{proposed goal paths} for each actor using the centerlines of mapped lanes in the scene, and then predicts a categorical distribution over these goals along with one or more trajectories for each goal. 

Importantly, by leveraging set-based neural network architectures~\citep{zaheer2017deep}, our model is able to handle an arbitrary number of goals for each actor in the scene. Because of this characteristic, our model is \emph{map adaptive}. Specifically, we can construct a set of goals from any scene with any lane topology, as long as the area has been mapped. We can therefore naturally handle N-way (e.g., 3-way, 4-way, 5-way, 6-way) intersections and other unusual map geometries (e.g., roundabouts, curvy roads). 

Although human drivers usually respect mapped lane boundaries, it is also critically important for an autonomy system to be able to predict motion that deviates from this norm. In order to capture non-map-compliant behavior, we augment our \emph{goal-based} trajectory modes with one or more additional \emph{motion-based} trajectory modes which aim to extrapolate the actor's current motion into the future, and are designed to capture the driving behavior that is not covered by the set of goals that we generate. 
By reasoning about the set of trajectories that are plausible when considering both goal-based and motion-based behavior, 
our approach is able to accurately represent the full distribution over the future locations of each vehicle in the scene using a compact set of trajectory modes. 

\begin{figure}
    \centering
    \subcaptionbox{
    The actor is strictly following a right-turning goal path.
    \label{fig:intro:left}}{\includegraphics[width=0.46\textwidth,trim={.5cm .5cm .5cm .5cm},clip]{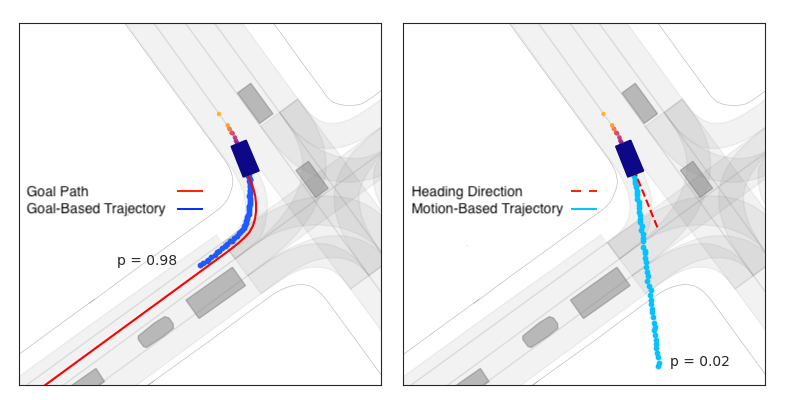}}
    \hspace{.5cm}
    \subcaptionbox{
    The actor is passing another actor ahead and not strictly following the proposed goal path.
    \label{fig:intro:right}}{\includegraphics[width=0.46\textwidth,trim={.5cm .5cm .5cm .5cm},clip]{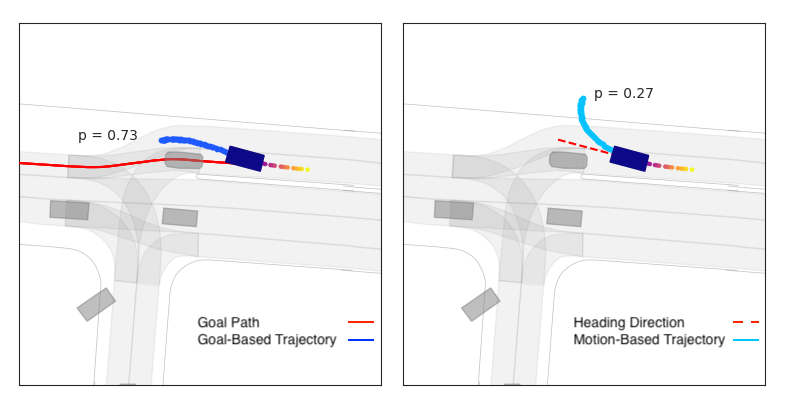}}
    \caption {Examples of two driving scenarios, (a) and (b), each with one actor of interest. In (a), the goal-based trajectory gets high probability (98\%) because the goal provides a good explanation for the actor's motion. In (b), since the candidate goal path cannot fully explain the actor's current motion, more probability mass (27\%) is assigned to the motion-based trajectory than in (a) (2\%). }
    \vspace{-.2cm}
    \label{fig:intro}
\end{figure}

Figure~\ref{fig:intro} highlights the duality of goal-based and motion-based trajectories. We show two different scenes, each with an actor of interest. In both cases, we just have a single proposed goal path for each actor, which means that the simplest variant of our model generates two trajectory predictions: one goal-based trajectory that uses the goal path as a spatial anchor, and one motion-based trajectory that uses the actor's heading as a reference direction. 
Together, these examples illustrate that our model is able to make use of the goal paths when they make sense, 
but also learns to fall back to the motion-based prediction when none of the goals can adequately explain the actor's current motion.

Overall, our model represents the driving scene for each actor as a composition of map elements (lane goals) and learns a representation of these elements along with a permutation-invariant function that aggregates information over the elements. 
By doing so, our approach is better able to generalize to out-of-distribution scenes than existing methods, some of which have the tendency to memorize the specific road and lane configurations observed during training. 
We verify this idea empirically by training and testing on driving data collected in two completely different cities, and show the results in Section~\ref{sec:method}. Combining these results with our intuition, we conclude that the compositional design of our model provides inductive bias that enables it to adapt to the local road geometry of any scene and still generate high-quality predictions even when that specific lane configuration has never before been seen by the model.

\section{Related Work}
\vspace{-.1cm}

Vehicle future motion forecasting has been an active area of research for the last few decades. Early works such as \citep{kaempchen2004imm, barth2008will, lytrivis2008cooperative} use physics-based kinematic models to predict future positions. Although this class of methods can be accurate in the short-term, they cannot capture long-term behavior of vehicles, which is mainly driven by the surrounding context and actors' intrinsic goals. Another line of research includes deep-learning-based methods such as \citep{deo2018convolutional}, which improve on the physics-based models by using recurrent networks to encode actor history and capture context from nearby actors. Recent work has focused largely on multi-modal prediction, where existing methods broadly fall into two categories: generative and discriminative. We discuss these in more detail below.

Generative models use stochastic sampling to approximate the distribution over future behavior. R2P2~\citep{rhinehart2018r2p2} generates trajectories using a one-step stochastic policy. DESIRE~\citep{lee2017desire} draws samples via a conditional variational autoencoder. Social GAN~\citep{gupta2018social} and SoPhie~\citep{sadeghian2019sophie} utilize GAN architectures to generate diverse and realistic samples. Overall, these sampling-based methods are challenging because a very large number of samples may be necessary in order to fully cover the distribution over future behavior, including interesting but low-probability regions of the distribution.

Discriminative models directly regress future trajectories and often include a classification loss to induce the model to select the trajectory mode that most closely matches the ground truth. A number of approaches have been proposed that use different methods to produce a diverse and useful set of discrete trajectory modes, which are often components in a mixture distribution. 
MTP~\citep{cui2019multimodal} learns multiple different trajectory modes via an unsupervised approach. MultiPath~\citep{chai2020multipath} predicts trajectory residuals from a fixed set of trajectory anchors learned using $k$-means. CoverNet~\citep{phan2020covernet} generates a large number of dynamically feasible trajectories via a set coverage approach. In this work, because we use a discriminative approach, we focus on comparing to other methods that do the same. 

Several trajectory prediction approaches have proposed different methods for capturing rich scene context in their input representations.
RasterNet~\citep{djuric2020uncertainty,chou2019predicting} rasterizes high-definition maps and the states of surrounding actors into a bird's eye view image. VectorNet~\citep{gao2020vectornet} directly uses vectorized map information and actor trajectories with minimal information loss. Other very recent works such as \citep{pan2019lane,kim2020multi,luo2020probabilistic} encode lane features using attention mechanisms and pursue some of the same themes that we investigate in this work. There is also a broad class of methods that jointly perform object detection and motion forecasting~\cite{luo2018fast,casas2018intentnet,djuric2020multinet}, which allows the model to incorporate rich scene context and appearance features that are learned in the perception stage into the prediction stage.

%

Our work relies heavily on the use of the Frenet-Serret path-relative coordinate frame~\citep{frenet1852courbes,serret1851quelques}, which is deeply rooted in differential geometry and is quite widely used for trajectory planning~\citep{kant1986toward, werling2010optimal}, though much less frequently employed for trajectory prediction. 
Our work also makes use of recent ideas for set-based \citep{zaheer2017deep} and graph-based \citep{scarselli2008graph, battaglia2018relational} neural network architectures which allow us to learn permutation-invariant and permutation-equivariant functions of an input set of elements.
The key differences between our method and previous methods lie in how we utilize structured map information: (1) both our model inputs and outputs are represented in the path-relative coordinate frame; (2) our model outputs a variable number of trajectories because it adapts to the map geometry.

\section{Method}
\label{sec:method}
\vspace{-.1cm}

\begin{figure}
    \centering
    \vspace{-.2cm}
    \includegraphics[width=\linewidth,trim={0cm .3cm 0cm .4cm},clip]{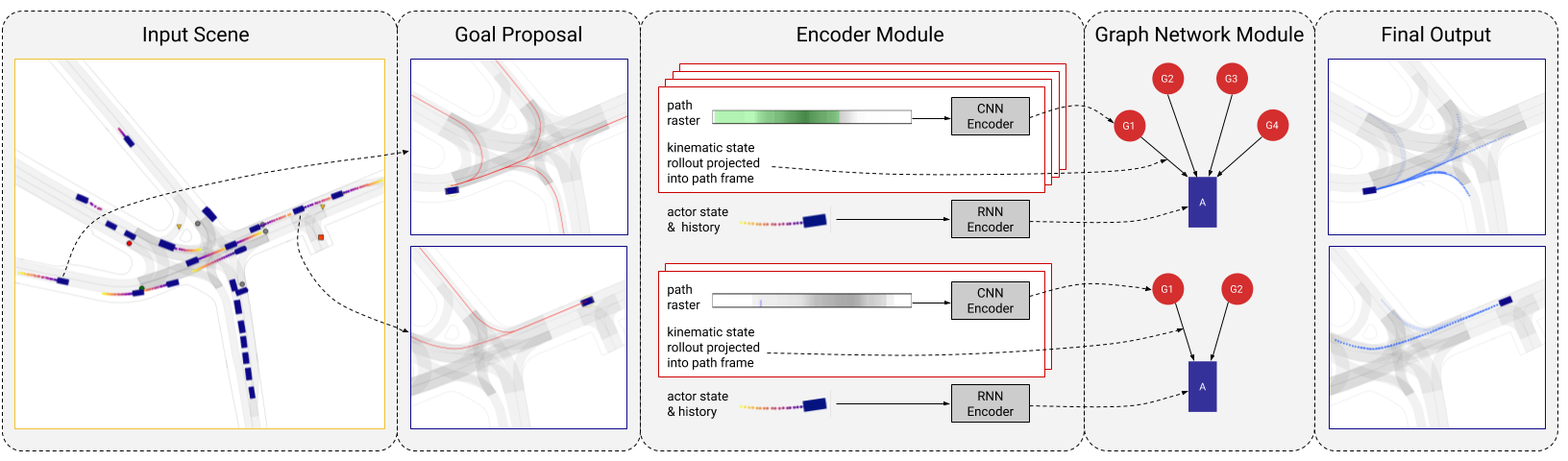}
    \caption{An overview of our multi-modal goal-based trajectory prediction method. 
    Given a scene, we process each actor in 3 stages: (a) generate a set of goal paths; (b) encode input representations for the actor states and goal paths; (c) predict a trajectory distribution using our graph network.
    }
    \vspace{-.1cm}
    \label{fig:method}
\end{figure}

In this section, we describe our novel method for goal-directed trajectory prediction, which we call GoalNet. For the purposes of this work, we assume that object detection and tracking have already been performed using sensor inputs from the scene. We are therefore presented with a set of detected objects with positions and headings $(x, y, \theta) $, bounding box dimensions $(w, h)$, higher-order states (velocity, acceleration), and 2 seconds of motion history (past $x,y$ positions). Given this information, we focus on predicting the future motion of all dynamic (non-parked) vehicles in the scene. A high-level overview of our method is provided in Figure \ref{fig:method} with three key components of our methods highlighted: (1) \emph{Goal Proposal}: We propose a set of goals for each actor based on local map geometry.
(2) \emph{Encoder Module}: The actor state and scene context are encoded in this module. (3) \emph{Graph Network Module}: We use a graph network to make predictions based on the encoded features. 
In the remainder of this section, we describe each module in detail.

\subsection{Goal Proposal}
\label{sec:method-goal-proposal}
\vspace{-.2cm}

The first step of our trajectory prediction method is to generate a set of candidate goals for each actor. To do this, we leverage high-definition maps of the scene, which are readily available in many self-driving systems. For the purposes of vehicle motion prediction, we define a \emph{goal} as a destination point along a mapped lane together with the spatial path to that destination. We generate a series of \emph{goal paths} for each actor using the centerlines of mapped lanes as the spatial paths. 

Specifically, given the bounding box of a detected vehicle, we first compute its centroid $c_x, c_y$, and then use a search radius $r$ around the centroid to identify all nearby starting lanes. We then follow the mapped lane sequences out to a fixed distance $d$ in order to construct a lane graph using the identified starting lanes as the root nodes. Finally, we generate the set of all paths from any root node to a leaf node in the lane graph to produce our set of goal paths. Each path is represented as a sequence of 2D points in space with no temporal component. In this work, we use the generated goal paths as spatial anchors for our trajectory predictions. 

A key feature of our approach is that 
we are able to generate a \emph{variable number of goals} for each actor, allowing our model to adapt to different road geometries at run time. 
 All legally plausible goal paths will be generated for an actor based on the map structure and the specified search radius $r$. 
For example, in Figure \ref{fig:method}, four goals are proposed for the actor approaching the intersection from the lower left corner. In contrast, the actor approaching from the top right corner only has two proposed goals because it is driving in a right-turn-only lane.


\subsection{Path-Relative Coordinate Frame}
\vspace{-.2cm}

\begin{figure}
    \centering
    \includegraphics[width=8.5cm]{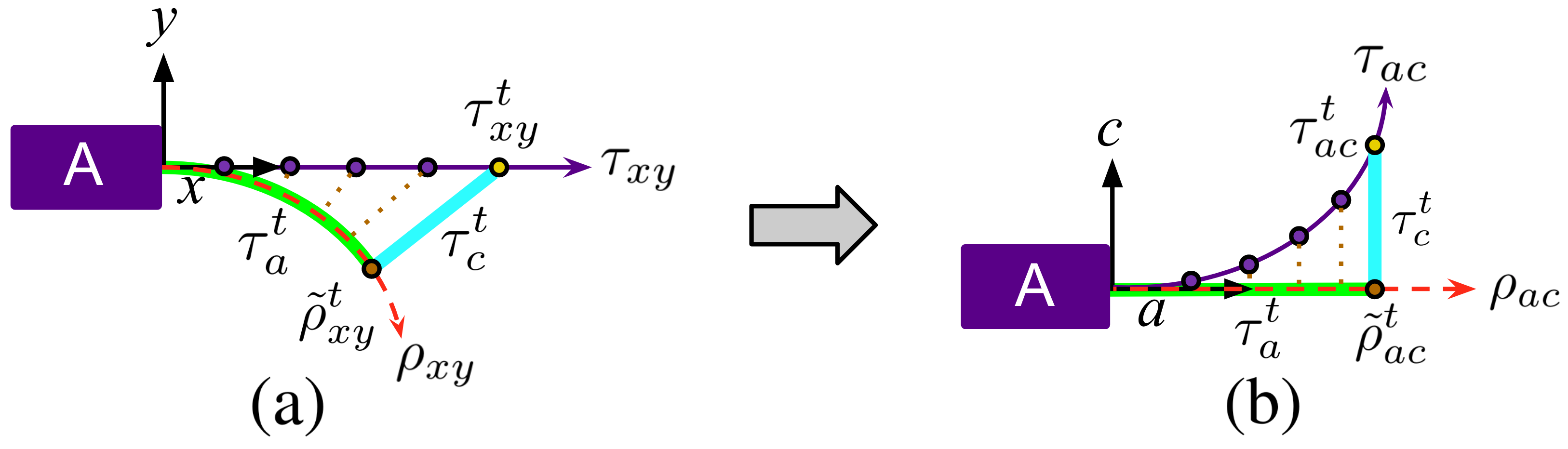} 
    \caption{An illustration of the transformation from the Cartesian $(x,y)$ coordinate frame (a) to the path-relative $(a,c)$ coordinate frame (b). Full details are provided in 
    Equations \ref{eqn:along-track-transform} and \ref{eqn:cross-track-transform}.} 
    \label{fig:path_relative}
    \vspace{-.1cm}
\end{figure}



In this section, we describe the path-relative coordinate frame that we use 
extensively in our model. 
We utilize the Frenet-Serret frame, defined in a piecewise linear manner along the ordinarily non-differentiable polyline.
Let $\rho_{xy} = (\rho_{xy}^{1}, \cdots, \rho_{xy}^{L})$ be a spatial reference path with 2D points $\rho_{xy}^{i} = (\rho_x^{i}, \rho_y^{i})$ evenly sampled along the path at a fixed spatial resolution $\delta_\rho$. Let $\tau_{xy} = (\tau_{xy}^{1}, \cdots, \tau_{xy}^{T})$ be a spatio-temporal trajectory with 2D waypoints $\tau_{xy}^{t} = (\tau_x^{t}, \tau_y^{t})$ evenly sampled in time. 

Given a path $\rho_{xy}$ and trajectory $\tau_{xy}$, whose points are both defined in the Cartesian coordinate frame, 
we define the along-track and cross-track decomposition for each trajectory waypoint as follows:
\begin{align}
     \tau_a^{t} &~=~ i^t \delta_\rho ~+~ \|\rho_{xy}^{i^t} - \tilde{\rho}_{xy}^{t} \|_2 
     \label{eqn:along-track-transform} \\
     \tau_c^{t} &~=~   \langle (\tau_{xy}^{t} - \tilde{\rho}_{xy}^{t}) ~,~ \xi (\rho_{xy}^{i^{t}+1} - \rho_{xy}^{i^t})_\perp \rangle \quad \text{where}~~ \xi = \|\rho_{xy}^{i^{t}+1} - \rho_{xy}^{i^t}\|_2^{-1} \label{eqn:cross-track-transform}
\end{align}
Here $\tilde{\rho}_{xy}^{t}$ denotes the closest point on the polyline of the path $\rho_{xy}$ to the trajectory point $\tau_{xy}^{t}$, $i^t$ denotes the index of the path point that precedes the closest point $\tilde{\rho}_{xy}^{t}$, and $\perp$ denotes orthogonal vector operation. When the closest point is at the end of the reference path, we assume $i^t$ to be $L-1$. Figure \ref{fig:path_relative} shows a graphical depiction of this coordinate frame transformation.

Finally, we define the path projection operator $\Pi_{\rho}(\cdot)$ as an operator that maps the Cartesian frame representation of a trajectory $\tau$ to its representation in the path-relative frame of a specific path $\rho$. In the remainder of this paper, when it is clear from context, we drop the subscripts $_{xy}$ and $_{ac}$. 

\subsection{Input and Output Representations}
\vspace{-.2cm}
\label{sec:method-input-output-representation}

We make heavy use of the path-relative coordinate frame for both our input representation and our output representation, which are described in detail below.

\textbf{Input Representation}: The inputs to our model consist of current and past actor states along with rich map context in the form of goal proposals. To encode this information into a useful form, we develop an Encoder Module (see Figure~\ref{fig:method}) with three parts: (1) \emph{Actor State}: The historical positions and current higher-order motion states for the actor of interest are encoded by an RNN and MLP encoder, respectively. (2) \emph{Along-Path Rasters}: We construct a path-aligned raster to capture the scene context along each goal path. The raster has multiple channels, which capture the path curvature, locations of traffic control regions, and the position and speeds of the closest surrounding actors to the actor of interest. Each channel has size 80 $\times$ 4, where each pixel covers a 1m $\times$ 1m area. This is encoded by a CNN encoder. (3) \emph{Path-Relative State Rollout}: 
We apply a kinematic equation to calculate the actor's extrapolated future positions from its current motion states according to $\tilde{s}_t = s_0 + v_0 \, t + 0.5 \, a_0 \, t^{2}$. Given this kinematic trajectory rollout $\tilde{s} = (\tilde{s}_0, \cdots, \tilde{s}_T)$, we project it to the path-relative frame of each goal path $\rho$ to obtain $e = \Pi_{\rho}(\tilde{s})$. Outputs from (1) and (2) are used as the actor node features $a$ and goal node features $g$ while (3) is used as the edge feature $e$ in the graph network as described in Section \ref{sec:method-graph-network}.

\textbf{Output Representation}: Given a reference goal path $\rho$, we predict a \emph{goal-based trajectory} for the goal by representing the trajectory in the path-relative frame. Specifically, instead of transforming the ground truth trajectory $\tau_{xy}$ to an actor-centric frame (centered at the actor's position and rotated to align with its heading) and directly regressing $\tau_{xy}$ as is done in typical existing approaches (e.g., \cite{cui2019multimodal}, \cite{gao2020vectornet}), we instead regress $\tau_{ac} = \Pi_{\rho}(\tau_{xy})$. This representation has many benefits, including: (1) it captures the lane topology in a frame of reference that allows the model to more easily learn to generalize across paths with different curvatures, and (2) it allows us to naturally decompose the spatial (cross-track) and temporal (along-track) dimensions of the actor's future motion.

\subsection{Spatial and Temporal Multi-Modality}
\label{sec:method-multi-modality}
\vspace{-.2cm}

Our model is multi-modal and contains two types of multi-modality: spatial and temporal. Let $N$ be the number of goal proposals for an actor and let $M$ be the number of temporal modes for each spatial mode. Our model outputs a total of $K=(N+1)\,M$ trajectories along with a categorical distribution over the $K$ modes. The ``ground truth'' mode probability is required for supervision in training and is assumed to be the product of (a) the target spatial mode probability and (b) the target temporal mode probability conditioned on its underlying spatial path. 

\textbf{Spatial Modes}: Our model contains two types of spatial modes: \emph{goal-based} and \emph{goal-free}. Each goal-based mode is associated with a goal. The target goal-based mode probability is computed by an algorithm that identifies goal-following behavior. An actor is considered to be following a goal if its future trajectory stays within a certain cross-track deviation from the goal path (see Appendix~\ref{app:path-autolabeling} for details). Because goal paths can partially overlap with one another and the actor's observed future trajectory might be quite short (e.g., consider a slow-moving actor), it is possible 
that multiple goal paths will match the actor's behavior. When there are $G > 0$ goals being followed, we assign equal target probability $1/G$ to each of these goals. 
To capture non-goal-following behaviors such as pulling onto the shoulder, our model also develops a goal-free spatial mode for each actor. %
The target probability for the goal-free mode 
is set to $1$ when none of the 
goals match the actor's 
future trajectory. Qualitatively, we observe that the goal-free mode typically predicts trajectories that extrapolate the actor's current motion, so we also refer to these as motion-based trajectories.

\textbf{Temporal Modes}: 
Within a spatial mode, an actor could exhibit different temporal behaviors, such as slowing down or speeding up. For each spatial mode, including both goal-based and goal-free modes, our model predicts a fixed number $M$ of temporal modes. Following previous work~\cite{cui2019multimodal}, we use an unsupervised approach to identify the ``ground truth'' temporal mode on the fly during training. Given the trajectories for the temporal modes of each 
spatial mode with nonzero target probability, the temporal mode whose trajectory is closest to the ground truth trajectory is selected as the ground truth temporal mode and assigned target probability 1. 







\subsection{Graph Network Formulation}
\label{sec:method-graph-network}
\vspace{-.2cm}

A central aspect of our approach is that each actor can have a variable number of candidate goal paths, where the number depends on the complexity of the lane geometry in the surrounding scene. In order to generate a corresponding number of trajectories, we draw inspiration from set-based and graph-based neural network architectures~\citep{zaheer2017deep,battaglia2018relational}. This style of architecture allows us to learn an equivariant function of the input set of goals, which is agnostic to the ordering of the goals. The function produces a corresponding set of output elements that have a one-to-one mapping with the input elements. In our case, we develop a very simple graph model that allows us to predict one or more trajectories per input goal along with a categorical distribution over the trajectories. 

The structure of our Graph Network Module is shown in Figure~\ref{fig:method}. Each graph contains two types of nodes: a single actor node $A$ and multiple goal nodes $G_j$. The graph also has directed edges $E_j$ that originate at each goal node and terminate at the actor node. Let $N$ be the number of goal nodes, $g_j = \phi_g(R_j)$ be the encoded path raster representation (goal node attributes), $a = \phi_s(H)$ be the encoded actor state representation (actor node attributes), and $e_j$ be the path-relative actor state rollout (edge attributes), as described in Section~\ref{sec:method-input-output-representation}. We execute two layers of graph network updates, using the following update functions applied in order from left to right:
\begin{align}
    \tilde{e}_j = \phi_e(a, e_j, g_j) \hspace{1.5cm}
    \bar{e} = \psi(\{ \tilde{e}_j \}_{j=1}^N) \hspace{1.5cm}
    \tilde{a} = \phi_a(a, \bar{e})
\end{align}
where $\tilde{e}_j$ and $\tilde{a}$ then become the edge and actor node representations that are used in the subsequent graph layer. We note that the goal node latent representation does not get updated in the graph network layers because it does not have any incoming edges in the graph. However, the edge latent representations eventually capture information from the full set of goals, not just one single goal. Here $\phi_e(\cdot)$ and $\phi_a(\cdot)$ are 2-layer MLPs and $\psi(\cdot)$ is a permutation-invariant aggregation function (we use the mean). Given the resulting node and edge representations, we predict the goal-based trajectories and probabilities from the actor-goal edges and predict the goal-free trajectories and probabilities from the actor node.


\subsection{Loss Function}
\label{sec:method-loss-function}
\vspace{-.2cm}

The output of our model consists of a set of goal-based trajectories, which are represented in the path-relative coordinate frame of their associated reference path, a set of goal-free trajectories, which are represented in the actor-centric coordinate frame, and the mode probabilities associated with each trajectory. For a given actor $i$, let $\tau^{(i)}$ denote its ground truth future trajectory, let $p^{(i, k)}$ denote the ground truth mode distribution over $K^{(i)}$ modes, let $\hat{\tau}^{(i,k)}$ denote the $k$-th predicted trajectory for the actor, and let $\hat{p}^{(i,k)}$ denote the predicted mode probability of the $k$-th predicted trajectory. 

Our overall loss function decomposes into a classification loss $\ell_{\textrm{cls}}$ and a regression loss $\ell_{\textrm{reg}}$, which we compute for each actor and sum over all actors. The two components of the loss are given by:
\begin{align}
    \ell_{\textrm{cls}}^{(i)} &= - \textstyle \sum_{k = 1}^{K^{(i)}} p^{(i,k)} \log{\hat{p}^{(i,k)}}\\
    \ell_{\textrm{reg}}^{(i)} &= \textstyle \sum_{k = 1}^{K^{(i)}} p^{(i, k)}  \left( \|\tau_a^{(i)} - \hat{\tau}_a^{(i,k)} \|_1 + \gamma \, \|\tau_c^{(i)} - \hat{\tau}_c^{(i,k)} \|_1 \right)
\end{align}
where $\gamma > 1$ is a scalar that up-weights the cross-track error relative to the along-track error (since the magnitude of the cross-track error is much smaller on average). Note that we do not treat the regression loss on goal-free modes differently. We treat the $x$-value and $y$-value of the goal-free trajectory in actor-centric (heading-aligned) frame to be along-track and cross-track, respectively.


\section{Experiments}
\vspace{-.1cm}


For our experiments, we train and evaluate two versions of our model: GoalNet-1T, which produces 1 temporal mode per spatial mode and thus only has spatial multi-modality, and GoalNet-2T, which produces 2 temporal modes per spatial mode. Implementation details can be found in Appendix~\ref{app:our-model-design}.


\subsection{Datasets}
\vspace{-.2cm}

In this work, we use two datasets for evaluation: our internal self-driving dataset and the public nuScenes dataset~\citep{nuscenes2019}. Our internal dataset has approximately 4 million frames collected from more than 100 hours of driving across two American cities. 
The nuScenes dataset has 700 train snippets and 150 validation snippets from Boston and Singapore. More details are given in Appendix \ref{app:perception}.

\subsection{Baselines}
\vspace{-.2cm}

All baselines perform multi-modal prediction and they all have similar input representation but have different output representations. Additional implementation details can be found in Appendix~\ref{app:baseline-implementations}.

\textbf{Multiple Trajectory Prediction}~\citep{cui2019multimodal}: This method learns trajectory modes in an unsupervised way. It directly regresses the future ground truth trajectory and only allows the gradient to flow back from the best-matching predicted trajectory. We use 3 modes for MTP.

\textbf{MultiPath}~\citep{chai2020multipath}: This method uses spatio-temporal trajectory anchors, which are estimated by running $k$-means on the training data. Its trajectory prediction classifies the ground truth anchor and regresses the offsets of the ground truth trajectory from this anchor. We use 64 modes for MultiPath.

\textbf{CoverNet}~\citep{phan2020covernet}: This method constructs a fixed or dynamic trajectory set that is ensured to provide $\epsilon$-coverage over all trajectories in the training set up to an $\epsilon$ tolerance. It then performs classification over the trajectory set with no additional regression component. 
Since CoverNet has a publicly released fixed trajectory set containing $2206$ trajectories for the nuScenes dataset, we directly use that in our nuScenes experiments. We do not experiment with the dynamic version of CoverNet.

\subsection{Results}
\vspace{-.2cm}

\begin{figure}
    \centering
    \begin{minipage}[t][][b]{.43\textwidth}
    \resizebox{\textwidth}{!}{
    \begin{tabular}{cc}
     \toprule
     Method & Number of Modes \\
     \midrule
       GoalNet-1T (internal) & 2.81 $\pm$1.91 \\
       GoalNet-1T (nuScenes) & 1.99 $\pm$1.46 \\
       MTP  & 3 \\
       MultiPath  & 64 \\
       CoverNet (nuScenes)  & 2206 \\
     \bottomrule
     \end{tabular}
     }
    \end{minipage}
    \captionof{table}{Comparison of the total number of trajectory modes produced by each method (we report mean $\pm$ stddev for GoalNet). Note that GoalNet-2T always has twice as many modes as GoalNet-1T.}
    \label{tab:compactness}
\end{figure}


We first compare the relative compactness of each method's multi-modal trajectory representation. Table \ref{tab:compactness} shows the number of trajectory modes used by each approach. GoalNet and MTP have a similar number of modes on average. Note that we have a large standard deviation, which provides further evidence that we adapt to different map geometries. By contrast, MultiPath and CoverNet require many more modes. This can have a significant impact on the run time of downstream motion planning systems, which often scale in the number of trajectories they must process. 


\begin{table}[tbp]
\begin{center}
\resizebox{\textwidth}{!}{
\begin{tabular}{lccccccc}
\toprule
Method & $\min_{1}$ADE  & $\min_{1}$FDE & $\min_{3}$ADE & $\min_{3}$FDE  & $\min_{5}$ADE  & $\min_{10}$ADE & $\mathbb{E}$[ADE] \\
\midrule
  MTP  & 2.67 (2.77) & 7.08 (7.65) & 1.69 (2.10) & 4.36 (5.92) & 1.69 (2.10) & 1.69 (2.10) & 2.98 (2.82) \\
  MultiPath  & 2.91 (4.01) & 7.69 (10.39) & 1.78 (2.40) & 4.61 (6.24) & 1.44 (1.85) & \textbf{1.14} (1.35) & 3.33 (4.13) \\
  CoverNet  & \, --- \, (4.57) & \, --- \, (10.37) & \, --- \, (2.99)  & \, --- \, (6.91) & \, --- \, (2.44) & \, --- \, (1.87) & \, --- \, (4.73) \\[1pt]
  \hdashline\\[-7pt]
  GoalNet-1T & \textbf{2.27} (\textbf{2.13}) & \textbf{5.90} (\textbf{5.79}) & 1.86 (1.77) & 4.70 (4.75) & 1.80 (1.75) & 1.79 (1.75) & \textbf{2.35} (\textbf{2.18}) \\
  GoalNet-2T  & 2.53 (2.44) & 6.57 (6.63) & \textbf{1.53} (\textbf{1.41}) & \textbf{3.83} (\textbf{3.66}) & \textbf{1.34} (\textbf{1.27}) & 1.28 (\textbf{1.22}) & 2.79 (2.60) \\
\bottomrule
\end{tabular}
}
\vspace{.2cm}
\caption{Trajectory prediction metrics on our internal dataset and on the public nuScenes dataset. Results listed as internal (nuScenes). All errors are reported in meters.}
\vspace{-.5cm}
\label{tab:results}
\end{center}
\end{table}

\begin{figure}
    \begin{minipage}[b][][b]{.48\textwidth}
    {\includegraphics[width=\linewidth,trim={.3cm 0cm .3cm .5cm},clip]{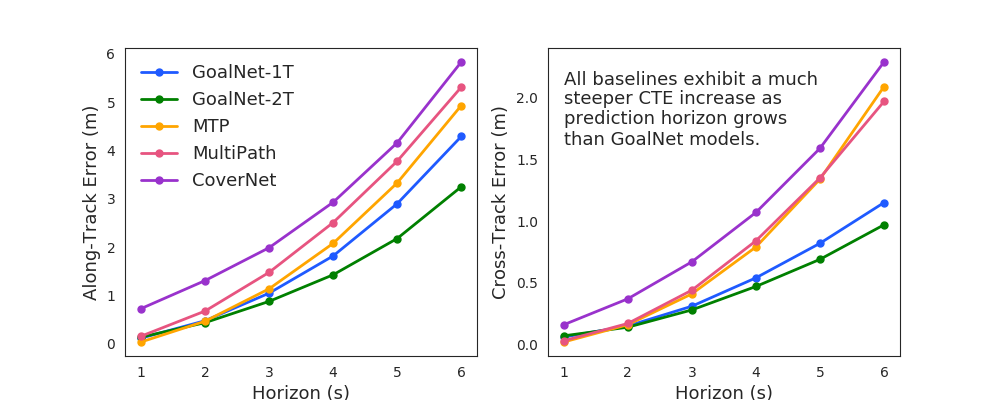}}
    \caption{A comparison of $\min_3$ along-track and cross-track errors at different time horizons.}
    \label{fig:at_ct}
    \end{minipage}
    \hfill
    \begin{minipage}[b][][b]{.48\textwidth}
    \resizebox{\textwidth}{!}{
    \begin{tabular}{c|cc}
    \hline
    \multirow{2}{*}{Method} & City 1 & City 2 \\
    & (in distribution) & (out of distribution) \\
    \hline
    GoalNet-2T & 2.59 & 2.62 (\textbf{+1.16}\% )  \\
    MTP & 2.66 & 2.89 (+8.65\%) \\
    MultiPath & 2.86 & 3.3 (+15.38\%) \\
    \hline
    \end{tabular}
    }
    \captionof{table}{Results of our inter-city generalization experiment. All models are trained only on data from City 1, and we report $\min_1$ADE on test sets from both City 1 and City 2.}
    \label{tab:generalization}
    \end{minipage}
    \vspace{-.2cm}
\end{figure}

In Table \ref{tab:results}, we compare our method with all baselines and demonstrate that we achieve the best performance in nearly all metrics on both our internal dataset and on the public nuScenes dataset. Similar to prior work, we report both the average displacement error over all horizons (ADE) and the final displacement error at 6s (FDE). We measure the min$_k$ trajectory error on the trajectory that best matches the ground truth (in terms of ADE) out of the $k$ most probable trajectories. We also report the expected (probability-weighted) average displacement error $\mathbb{E}$[ADE].\footnote{When calculating expected error for CoverNet, for computational efficiency, we keep only the 100 most probable trajectory modes and renormalize their probabilities.} GoalNet-1T has the smallest trajectory error when evaluating the most probable mode ($k=1$) and the expected error. GoalNet-2T has the best performance on nearly all other metrics. In addition to these displacement error metrics, we also evaluate the along-track error (ATE) and cross-track error (CTE), which we define in Appendix~\ref{app:metrics}. We plot ATE and CTE on nuScenes as a function of the prediction horizon in Figure~\ref{fig:at_ct}. From this plot, we see that GoalNet-2T outperforms all baselines across all horizons on both metrics. However, more notably, both versions of our model achieve a very substantial (40-50\%) gain in cross-track error at the longer time horizons. Furthermore, the gap between GoalNet and baseline methods quickly widens as the horizon increases, which demonstrates its benefits for longer-term prediction and underscores the utility of our map-adaptive spatial goals.


\begin{table}[tbp]
\begin{center}
\resizebox{\textwidth}{!}{
\begin{tabular}{ccccccccccc}
\toprule
\multirow{2}{1.3cm}{\centering path \\ rasters} & \multirow{2}{2.1cm}{\centering path edge features} & \multirow{2}{1.8cm}{\centering path-relative frame} & \multicolumn{4}{c}{$\min_{1}$} & \multicolumn{4}{c}{$\min_{3}$}  \\
&&& AATE & ACTE &  ADE &  FDE & AATE &  ACTE & ADE & FDE \\ 
\cmidrule(lr){1-3}  \cmidrule(lr){4-7}  \cmidrule(lr){8-11}
\cmark & \cmark &  \cmark  & \textbf{2.79} & 1.14 & \textbf{3.44} & \textbf{9.29} & \textbf{1.77} & \textbf{0.74} & \textbf{2.15} & \textbf{5.67} \\
{\color{lightgray} \xmark} & \cmark & \cmark  & 3.04 & \textbf{1.06} & 3.62 & 9.84 & 2.02 & \textbf{0.74} & 2.39 & 6.36 \\
\cmark & {\color{lightgray} \xmark} & \cmark & 3.28 & 1.42 & 4.04 & 10.37 & 1.95 & 0.88 & 2.40 & 6.15 \\
\cmark & \cmark & {\color{lightgray} \xmark} & 2.89 & 1.35 & 3.66 & 10.10 & 1.88 & 0.98 & 2.40 & 6.64 \\ 
\bottomrule
\end{tabular}
}
\vspace{.2cm}
\caption{Ablation study of GoalNet. Results are reported on our internal dataset, and all variations are evaluated only on challenging turning samples (which constitute 21.7\% of the data). Here AATE and ACTE are the average along-track and cross-track error (averaged over all horizons). 
}
\vspace{-.5cm}
\label{tab:ablation}
\end{center}
\end{table}

In Table \ref{tab:ablation}, we assess the importance of the different components of our model. For this study, we report results only on turning samples because for actors who are going straight, there will be no difference between using the path-relative coordinate frame and the actor-centric frame. When dropping the path raster, we see that ATE degrades, which suggests that knowledge of traffic signs, surrounding actor information, and path curvature helps the model to better reason about the temporal behavior of the actor. When dropping the path edge features, we see that both ATE and CTE are significantly degraded, which suggests that projecting the actor's kinematic state rollout into the path-relative frame eases the learning process. Lastly, representing both input scene and output trajectories in the path-relative frame helps reason about actor future behavior around intersections.

Next, we conduct an experiment to determine whether our model is able to generalize to unseen driving scenarios. Our internal dataset contains driving data from two different cities. We observe that City 2 has more unusual and challenging road geometries than City 1, so we train all models on data from City 1 but evaluate them on City 2. As shown in Table~\ref{tab:generalization}, all models have regressed performance when switching evaluation from City 1 to City 2. However, ours exhibits the least regression, which provides evidence that the map-adaptive and compositional design of our model allows it to generalize quite well to out-of-distribution samples. 

\begin{figure}[tbp]
\hspace{0.4cm}
\begin{minipage}{\textwidth}
    \begin{minipage}{0.12\textwidth}
    \centering
    \small
    5-way intersection
    \end{minipage}
    \begin{minipage}{0.72\textwidth}
    \centering
    {\includegraphics[width=.24\linewidth,trim={.75cm .5cm 21.5cm .6cm},clip]{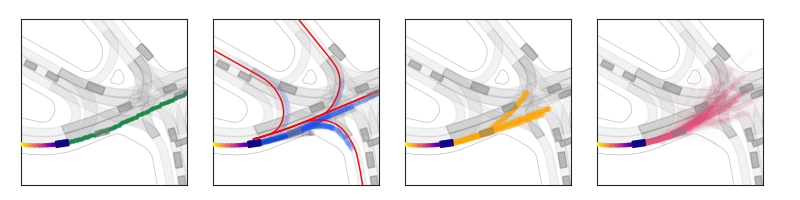}}
    {\includegraphics[width=.24\linewidth,trim={7.5cm .5cm 14.75cm .6cm},clip]{figures/qualitative_example_5_way_intersection.png}}        {\includegraphics[width=.24\linewidth,trim={14.25cm .5cm 8.0cm .6cm},clip]{figures/qualitative_example_5_way_intersection.png}}     {\includegraphics[width=.24\linewidth,trim={21cm .5cm 1.25cm .6cm},clip]{figures/qualitative_example_5_way_intersection.png}}
    \end{minipage}
    \\ 
    \begin{minipage}{0.12\textwidth}
    \centering
    \small
    curved\\ road
    \end{minipage}
    \begin{minipage}{0.72\textwidth}
    \centering
    {\includegraphics[width=.24\linewidth,trim={.75cm .5cm 21.5cm .5cm},clip]{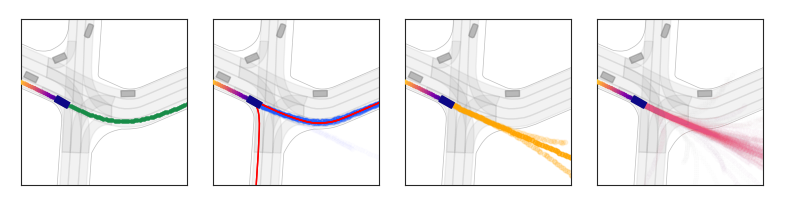}}
    {\includegraphics[width=.24\linewidth,trim={7.5cm .5cm 14.75cm .5cm},clip]{figures/qualitative_example_curved_road.png}} {\includegraphics[width=.24\linewidth,trim={14.25cm .5cm 8.0cm .5cm},clip]{figures/qualitative_example_curved_road.png}} {\includegraphics[width=.24\linewidth,trim={21cm .5cm 1.25cm .5cm},clip]{figures/qualitative_example_curved_road.png}}
    \end{minipage}
    \\
    \begin{minipage}{0.12\textwidth}
    \centering
    \small
    off-map driving
    \end{minipage}
    \begin{minipage}{0.72\textwidth}
    \centering
    \subcaptionbox{Truth\label{fig:qualitative:a}}{\includegraphics[width=.24\linewidth,trim={.75cm .5cm 21.5cm .5cm},clip]{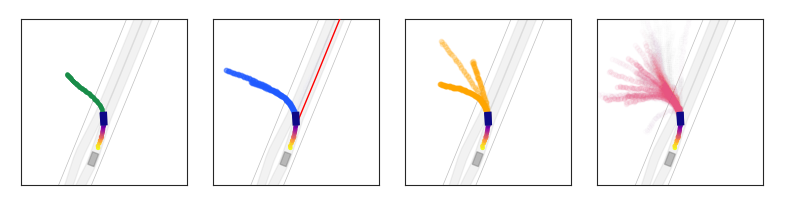}}
    \subcaptionbox{GoalNet\label{fig:qualitative:b}}{\includegraphics[width=.24\linewidth,trim={7.5cm .5cm 14.75cm .5cm},clip]{figures/qualitative_example_off_map.png}}
    \subcaptionbox{MTP\label{fig:qualitative:c}}{\includegraphics[width=.24\linewidth,trim={14.25cm .5cm 8.0cm .5cm},clip]{figures/qualitative_example_off_map.png}}
    \subcaptionbox{MultiPath\label{fig:qualitative:d}}{\includegraphics[width=.24\linewidth,trim={21cm .5cm 1.25cm .5cm},clip]{figures/qualitative_example_off_map.png}}
    \end{minipage}
\end{minipage}
\caption{Qualitative examples from our data. The left column shows the actor-of-interest's ground truth future trajectory (green), and the subsequent columns show the predicted trajectories from GoalNet (blue), MTP (yellow), and MultiPath (pink). For all methods, the trajectory probability is encoded in the alpha opacity value. For GoalNet (our method), we also show the goal paths in red.}
    \label{fig:qualitative}
    \vspace{-.3cm}
\end{figure}

In Figure \ref{fig:qualitative}, we showcase the ability of our method to handle both map-compliant driving behavior in unusual map geometries and non-map-compliant driving behavior. In the first example of a 5-way intersection, we are able to assign non-negligible probability to all modes and produce nice, lane-following trajectories. 
In contrast, the trajectory modes of MTP and MultiPath are much less spatially diverse. 
In the second example, our predicted trajectory is able to follow the curved road very well and match the ground truth behavior, whereas both MTP and MultiPath extrapolate the actor motion and fail to capture the challenging curved lane-following behavior. 
In the last example of off-map driving, although we do not have a goal that could capture the ground truth behavior, our motion-based trajectory still covers this mode by reasoning about the actor's current motion. 


\section{Conclusion}
\vspace{-.1cm}

In this work, we introduce GoalNet, a novel trajectory prediction model. We derive goals directly from the underlying map and use the reference paths to provide useful structure for both the model input and output representations. We produce a semantically interpretable probability distribution over the goal-based and motion-based modes. We demonstrate that 
we are able to achieve state-of-the-art performance and also adapt to unseen map configurations in new cities. 

\bibliography{references}

\begin{thebibliography}{28}
\providecommand{\natexlab}[1]{#1}
\providecommand{\url}[1]{\texttt{#1}}
\expandafter\ifx\csname urlstyle\endcsname\relax
  \providecommand{\doi}[1]{doi: #1}\else
  \providecommand{\doi}{doi: \begingroup \urlstyle{rm}\Url}\fi

\bibitem[Zaheer et~al.(2017)Zaheer, Kottur, Ravanbakhsh, Poczos, Salakhutdinov,
  and Smola]{zaheer2017deep}
M.~Zaheer, S.~Kottur, S.~Ravanbakhsh, B.~Poczos, R.~R. Salakhutdinov, and A.~J.
  Smola.
\newblock Deep sets.
\newblock In \emph{Advances in neural information processing systems}, pages
  3391--3401, 2017.

\bibitem[Kaempchen et~al.(2004)Kaempchen, Weiss, Schaefer, and
  Dietmayer]{kaempchen2004imm}
N.~Kaempchen, K.~Weiss, M.~Schaefer, and K.~C. Dietmayer.
\newblock {IMM} object tracking for high dynamic driving maneuvers.
\newblock In \emph{IEEE Intelligent Vehicles Symposium, 2004}, pages 825--830.
  IEEE, 2004.

\bibitem[Barth and Franke(2008)]{barth2008will}
A.~Barth and U.~Franke.
\newblock Where will the oncoming vehicle be the next second?
\newblock In \emph{2008 IEEE Intelligent Vehicles Symposium}, pages 1068--1073.
  IEEE, 2008.

\bibitem[Lytrivis et~al.(2008)Lytrivis, Thomaidis, and
  Amditis]{lytrivis2008cooperative}
P.~Lytrivis, G.~Thomaidis, and A.~Amditis.
\newblock Cooperative path prediction in vehicular environments.
\newblock In \emph{2008 11th International IEEE Conference on Intelligent
  Transportation Systems}, pages 803--808. IEEE, 2008.

\bibitem[Deo and Trivedi(2018)]{deo2018convolutional}
N.~Deo and M.~M. Trivedi.
\newblock Convolutional social pooling for vehicle trajectory prediction.
\newblock In \emph{Proceedings of the IEEE Conference on Computer Vision and
  Pattern Recognition Workshops}, pages 1468--1476, 2018.

\bibitem[Rhinehart et~al.(2018)Rhinehart, Kitani, and
  Vernaza]{rhinehart2018r2p2}
N.~Rhinehart, K.~M. Kitani, and P.~Vernaza.
\newblock {R2P2}: A reparameterized pushforward policy for diverse, precise
  generative path forecasting.
\newblock In \emph{Proceedings of the European Conference on Computer Vision
  (ECCV)}, pages 772--788, 2018.

\bibitem[Lee et~al.(2017)Lee, Choi, Vernaza, Choy, Torr, and
  Chandraker]{lee2017desire}
N.~Lee, W.~Choi, P.~Vernaza, C.~B. Choy, P.~H. Torr, and M.~Chandraker.
\newblock {DESIRE}: Distant future prediction in dynamic scenes with
  interacting agents.
\newblock In \emph{Proceedings of the IEEE Conference on Computer Vision and
  Pattern Recognition}, pages 336--345, 2017.

\bibitem[Gupta et~al.(2018)Gupta, Johnson, Fei-Fei, Savarese, and
  Alahi]{gupta2018social}
A.~Gupta, J.~Johnson, L.~Fei-Fei, S.~Savarese, and A.~Alahi.
\newblock Social {GAN}: Socially acceptable trajectories with generative
  adversarial networks.
\newblock In \emph{Proceedings of the IEEE Conference on Computer Vision and
  Pattern Recognition}, pages 2255--2264, 2018.

\bibitem[Sadeghian et~al.(2019)Sadeghian, Kosaraju, Sadeghian, Hirose,
  Rezatofighi, and Savarese]{sadeghian2019sophie}
A.~Sadeghian, V.~Kosaraju, A.~Sadeghian, N.~Hirose, H.~Rezatofighi, and
  S.~Savarese.
\newblock So{P}hie: An attentive {GAN} for predicting paths compliant to social
  and physical constraints.
\newblock In \emph{Proceedings of the IEEE Conference on Computer Vision and
  Pattern Recognition}, pages 1349--1358, 2019.

\bibitem[Cui et~al.(2019)Cui, Radosavljevic, Chou, Lin, Nguyen, Huang,
  Schneider, and Djuric]{cui2019multimodal}
H.~Cui, V.~Radosavljevic, F.-C. Chou, T.-H. Lin, T.~Nguyen, T.-K. Huang,
  J.~Schneider, and N.~Djuric.
\newblock Multimodal trajectory predictions for autonomous driving using deep
  convolutional networks.
\newblock In \emph{2019 International Conference on Robotics and Automation
  (ICRA)}, pages 2090--2096. IEEE, 2019.

\bibitem[Chai et~al.(2020)Chai, Sapp, Bansal, and Anguelov]{chai2020multipath}
Y.~Chai, B.~Sapp, M.~Bansal, and D.~Anguelov.
\newblock {MultiPath}: Multiple probabilistic anchor trajectory hypotheses for
  behavior prediction.
\newblock In \emph{Conference on Robot Learning}, pages 86--99, 2020.

\bibitem[Phan-Minh et~al.(2020)Phan-Minh, Grigore, Boulton, Beijbom, and
  Wolff]{phan2020covernet}
T.~Phan-Minh, E.~C. Grigore, F.~A. Boulton, O.~Beijbom, and E.~M. Wolff.
\newblock {CoverNet}: Multimodal behavior prediction using trajectory sets.
\newblock In \emph{Proceedings of the IEEE/CVF Conference on Computer Vision
  and Pattern Recognition}, pages 14074--14083, 2020.

\bibitem[Djuric et~al.(2020)Djuric, Radosavljevic, Cui, Nguyen, Chou, Lin,
  Singh, and Schneider]{djuric2020uncertainty}
N.~Djuric, V.~Radosavljevic, H.~Cui, T.~Nguyen, F.-C. Chou, T.-H. Lin,
  N.~Singh, and J.~Schneider.
\newblock Uncertainty-aware short-term motion prediction of traffic actors for
  autonomous driving.
\newblock In \emph{The IEEE Winter Conference on Applications of Computer
  Vision}, pages 2095--2104, 2020.

\bibitem[Chou et~al.(2019)Chou, Lin, Cui, Radosavljevic, Nguyen, Huang,
  Niedoba, Schneider, and Djuric]{chou2019predicting}
F.-C. Chou, T.-H. Lin, H.~Cui, V.~Radosavljevic, T.~Nguyen, T.-K. Huang,
  M.~Niedoba, J.~Schneider, and N.~Djuric.
\newblock Predicting motion of vulnerable road users using high-definition maps
  and efficient convnets.
\newblock \emph{arXiv preprint arXiv:1906.08469}, 2019.

\bibitem[Gao et~al.(2020)Gao, Sun, Zhao, Shen, Anguelov, Li, and
  Schmid]{gao2020vectornet}
J.~Gao, C.~Sun, H.~Zhao, Y.~Shen, D.~Anguelov, C.~Li, and C.~Schmid.
\newblock {VectorNet}: Encoding {HD} maps and agent dynamics from vectorized
  representation.
\newblock In \emph{Proceedings of the IEEE/CVF Conference on Computer Vision
  and Pattern Recognition}, pages 11525--11533, 2020.

\bibitem[Pan et~al.(2019)Pan, Sun, Xu, Jiang, Xiao, Hu, and Miao]{pan2019lane}
J.~Pan, H.~Sun, K.~Xu, Y.~Jiang, X.~Xiao, J.~Hu, and J.~Miao.
\newblock Lane attention: Predicting vehicles' moving trajectories by learning
  their attention over lanes.
\newblock \emph{arXiv preprint arXiv:1909.13377}, 2019.

\bibitem[Kim et~al.(2020)Kim, Kim, Kim, Cho, and Huh]{kim2020multi}
H.~Kim, D.~Kim, G.~Kim, J.~Cho, and K.~Huh.
\newblock Multi-head attention-based probabilistic vehicle trajectory
  prediction.
\newblock \emph{arXiv preprint arXiv:2004.03842}, 2020.

\bibitem[Luo et~al.(2020)Luo, Sun, Dabiri, and Yuille]{luo2020probabilistic}
C.~Luo, L.~Sun, D.~Dabiri, and A.~Yuille.
\newblock Probabilistic multi-modal trajectory prediction with lane attention
  for autonomous vehicles.
\newblock \emph{arXiv preprint arXiv:2007.02574}, 2020.

\bibitem[Luo et~al.(2018)Luo, Yang, and Urtasun]{luo2018fast}
W.~Luo, B.~Yang, and R.~Urtasun.
\newblock Fast and furious: Real time end-to-end {3D} detection, tracking and
  motion forecasting with a single convolutional net.
\newblock In \emph{Proceedings of the IEEE conference on Computer Vision and
  Pattern Recognition}, pages 3569--3577, 2018.

\bibitem[Casas et~al.(2018)Casas, Luo, and Urtasun]{casas2018intentnet}
S.~Casas, W.~Luo, and R.~Urtasun.
\newblock {IntentNet}: Learning to predict intention from raw sensor data.
\newblock In \emph{Conference on Robot Learning}, pages 947--956, 2018.

\bibitem[Djuric et~al.(2020)Djuric, Cui, Su, Wu, Wang, Chou, Martin, Feng, Hu,
  Xu, et~al.]{djuric2020multinet}
N.~Djuric, H.~Cui, Z.~Su, S.~Wu, H.~Wang, F.-C. Chou, L.~S. Martin, S.~Feng,
  R.~Hu, Y.~Xu, et~al.
\newblock {MultiNet}: Multiclass multistage multimodal motion prediction.
\newblock \emph{arXiv preprint arXiv:2006.02000}, 2020.

\bibitem[Fr{\'e}net(1852)]{frenet1852courbes}
F.~Fr{\'e}net.
\newblock Sur les courbes {\`a} double courbure.
\newblock \emph{Journal de math{\'e}matiques pures et appliqu{\'e}es}, pages
  437--447, 1852.

\bibitem[Serret(1851)]{serret1851quelques}
J.-A. Serret.
\newblock Sur quelques formules relatives {\`a} la th{\'e}orie des courbes
  {\`a} double courbure.
\newblock \emph{Journal de math{\'e}matiques pures et appliqu{\'e}es}, pages
  193--207, 1851.

\bibitem[Kant and Zucker(1986)]{kant1986toward}
K.~Kant and S.~W. Zucker.
\newblock Toward efficient trajectory planning: The path-velocity
  decomposition.
\newblock \emph{The international journal of robotics research}, 5\penalty0
  (3):\penalty0 72--89, 1986.

\bibitem[Werling et~al.(2010)Werling, Ziegler, Kammel, and
  Thrun]{werling2010optimal}
M.~Werling, J.~Ziegler, S.~Kammel, and S.~Thrun.
\newblock Optimal trajectory generation for dynamic street scenarios in a
  {F}r{\'e}net frame.
\newblock In \emph{2010 IEEE International Conference on Robotics and
  Automation}, pages 987--993. IEEE, 2010.

\bibitem[Scarselli et~al.(2008)Scarselli, Gori, Tsoi, Hagenbuchner, and
  Monfardini]{scarselli2008graph}
F.~Scarselli, M.~Gori, A.~C. Tsoi, M.~Hagenbuchner, and G.~Monfardini.
\newblock The graph neural network model.
\newblock \emph{IEEE Transactions on Neural Networks}, 20\penalty0
  (1):\penalty0 61--80, 2008.

\bibitem[Battaglia et~al.(2018)Battaglia, Hamrick, Bapst, Sanchez-Gonzalez,
  Zambaldi, Malinowski, Tacchetti, Raposo, Santoro, Faulkner,
  et~al.]{battaglia2018relational}
P.~W. Battaglia, J.~B. Hamrick, V.~Bapst, A.~Sanchez-Gonzalez, V.~Zambaldi,
  M.~Malinowski, A.~Tacchetti, D.~Raposo, A.~Santoro, R.~Faulkner, et~al.
\newblock Relational inductive biases, deep learning, and graph networks.
\newblock \emph{arXiv preprint arXiv:1806.01261}, 2018.

\bibitem[Caesar et~al.(2019)Caesar, Bankiti, Lang, Vora, Liong, Xu, Krishnan,
  Pan, Baldan, and Beijbom]{nuscenes2019}
H.~Caesar, V.~Bankiti, A.~H. Lang, S.~Vora, V.~E. Liong, Q.~Xu, A.~Krishnan,
  Y.~Pan, G.~Baldan, and O.~Beijbom.
\newblock {nuScenes}: A multimodal dataset for autonomous driving.
\newblock \emph{arXiv preprint arXiv:1903.11027}, 2019.

\end{thebibliography}

\appendix
\BeforeBeginEnvironment{appendices}{\clearpage}

\begin{appendices}

\section{Method Design Details}
\label{app:implementation}

\subsection{Path Auto-Labeling Algorithm}
\label{app:path-autolabeling}
\vspace{-.1cm}

The following section provides a detailed description of the path autolabeling algorithm that was introduced in Section \ref{sec:method-multi-modality}. In particular, given the ground truth trajectory $\tau_{xy}$ of the actor of interest and their $N$ candidate paths $\rho^{1}_{xy}, \rho^{2}_{xy}, \cdots , \rho^{N}_{xy}$, the path auto-labeling algorithm identifies zero or more paths taken by the actor. 

The algorithm is done in three steps. First, for each path $n$, we compute maximum cross-track deviation $\tau_{c_{\max}}^{n}$ of the trajectory $\tau_{xy}$ relative to the path $\rho^{n}_{xy}$. Specifically, 
\begin{align*}
\tau_{ac} &= \Pi_{\rho^{n}}(\tau_{xy}) \\
\tau_{c_{\max}}^{n} &= \max_{t \in \{1, \cdots, T \}} \tau_{c}^t
\end{align*}
Second, among all paths, we choose the path with the minimum value of the maximum cross-track deviation, namely: $\tau^m = \argmin (\tau_{c_{\max}}^{n})$. We then select all other paths whose maximum cross-track deviation is within a tolerance, which we set to 0.1m, of that of $\tau^m$. Finally, if $\tau^m_c < 5.0$m, we label all chosen paths as being followed. Otherwise, no path is followed. 

\subsection{Implementation Details}
\label{app:our-model-design}
\vspace{-.1cm}

The following section provides implementation details of the various model components described in Section \ref{sec:method}. 

\textbf{Path Generation}: In our path generation algorithm, we use a search radius of $r = 2$ meters based on the fact that the average lane is mostly between 3 meters and 4 meters wide. We use a fixed path length of $d = 80$ meters because a vehicle traveling at a typical city speed of $30$mph will travel 80 meters in 6 seconds. 
Finally, we use a constant sampling interval of 1 meter for the path points. 

\textbf{Model Architecture}: Our path-relative rasters are discretized to have shape (80, 4) with a resolution of 1m $\times$ 1m, which corresponds to 80 meters in along-path distance and 4 meters in cross-path distance. The path is placed at the central line of the raster. Our rasters have 8 channels in total: one channel captures the path curvature; two channels capture the position and speed of the 20 closest actors to the target actor; the remaining channels encode traffic sign and signal information, including stop signs, yield signs, green traffic lights, red traffic lights, and all other traffic lights. We encode each path raster using a CNN encoder. The overall raster shape is (80, 4, 8). We apply three 2D convolutional layers on top of each path raster with kernel size (3, 1), and then apply a 2D max pooling layer to get the encoded goal node features. We use a RNN encoder for the 2-second actor history sampled at 10Hz and a MLP encoder for its current estimated state to form the actor node features. The path-relative future roll-out is generated at 2Hz for a duration of 6 seconds and is directly fed into the graph network as edge features. 

\textbf{Model Prediction}: Given the node and edge representations that result from the graph network updates, we generate both edge-level and node-level predictions. From each edge, we output the goal-based spatial score, the associated temporal scores conditioned on the specific goal-based mode, and the goal-based trajectories (one for each temporal mode). From the actor node, we output the goal-free spatial score, the associated temporal scores conditioned on the goal-free mode, and the goal-free trajectories (one for each temporal mode). 
Finally, the spatial mode distribution is computed by taking a softmax over the spatial scores and the conditional temporal mode distributions are computed by taking a softmax over the temporal scores of each spatial mode. We end up with a joint categorical distribution over all goal-based and goal-free trajectories, where each trajectory probability is the product of its spatial and temporal probabilities.

\section{Experimental Settings}
\label{app:experiments}

\subsection{Datasets}
\label{app:perception}
\vspace{-.1cm}

The ground truth annotations have a frequency of 10Hz in our internal dataset and 2Hz in nuScenes. Since we don't have access to the ground truth labels for the official nuScenes test set, we randomly subsample 5\% of the logs from the train snippets to form our train-val dataset, and use the official validation snippets as our test set. For training, we exclude actors that are parked, actors that do not have at least 6 seconds of future observations, and the ego-vehicle itself. For evaluation, we additionally exclude all actors whose future trajectories move less than 1 meter over the 6-second horizon (these are generally easier cases).

Our model does not operate directly on sensor data but on detected and tracked objects. To obtain the object detections and state estimates for our internal dataset, we run an upstream perception module and use human-labeled data for the ground truth future positions. For the nuScenes dataset, we use human labels for both the actor state estimates and ground truth future positions.

\subsection{Baseline Implementations}
\label{app:baseline-implementations}
\vspace{-.1cm}
For implementations of different baselines, we use the same configuration for input representation to emphasize the difference in their output representation. In particular, when constructing the scene level raster images, we rotate the scene to align with the heading of the sensing vehicle. We have a full 2-second history channel sampled at 10Hz for each actor in the current timestamp, 8 rasterized map channels which captures different map semantics, and 3 different traffic sign channels including yield sign, stop sign and traffic signals. Our scene image has shape (640, 960) which corresponds to an area of size 100m $\times$ 150m. We first apply ResNet50 with a depth multiplier 25\% on the scene image which is downscaled to have size (160, 240). For each actor, we crop an actor-centric image with a patch size (64, 64) corresponding to an area of size 40m $\times$ 40m. An actor-centric convolution operation is then applied on top of the cropped image. The final pooled feature is concatenated together with the actor's velocity and acceleration and fed into an MLP to generate the final prediction. 

For CoverNet specifically, since the method partially depends on having a fixed set of pre-generated trajectories, we do not report CoverNet results on our internal dataset. For nuScenes, we make use of the publicly available CoverNet trajectory set. 

\section{Metrics Definitions}
\label{app:metrics}

\subsection{Displacement Error}
\vspace{-.1cm}
To evaluate a multi-modal set of predicted trajectories, we take the best trajectory (in terms of minimum average displacement) among the top-$k$ highest-probability trajectories. We refer to this selection as $\min_k$ in our evaluation. Note that $\min_1$ selects the most probable trajectory. Given a selected predicted trajectory $\hat{\tau}_{xy}$ and ground truth trajectory $\tau_{xy}$, we compute the average displacement error over the full horizon (ADE) 
and the final displacement error at 6 seconds (FDE) 
as follows:
\begin{align*}
    \text{ADE} &= \textstyle \frac{1}{T}\sum_{t=1}^{T} \| \tau_{xy}^t - \hat{\tau}_{xy}^t \|_2 \\
    \text{FDE} &= \| \tau_{xy}^T - \hat{\tau}_{xy}^T \|_2
\end{align*}
We also report the expected average displacement error, which is defined as:
\begin{align*}
    \mathbb{E}[\text{ADE}] = \textstyle \sum_{k=1}^K \hat{p}^k \text{ADE}(\hat{\tau}^k_{xy}, \tau_{xy})
\end{align*}
where $K$ is the total number of trajectory modes, $\hat{p}^k$ is the predicted probability of the $k$-th trajectory mode, $\hat{\tau}^k_{xy}$ is the $k$-th predicted trajectory, and $\tau_{xy}$ is the ground truth trajectory.


\subsection{Along-Track and Cross-Track Error}
\vspace{-.1cm}
The along-track error and cross-track error are a decomposition of the trajectory prediction error in the path-relative coordinate frame of the ground truth trajectory. Given the ground truth trajectory $\tau_{xy}$ and the predicted trajectory $\hat{\tau}_{xy}$, we first strip away the temporal component of $\tau_{xy}$ by re-sampling $\tau_{xy}$ at a fixed spatial resolution $\delta_\tau = 0.1$m to obtain ground truth path $\rho^*_{xy}$. Then, we project $\hat{\tau}_{xy}$ and $\tau_{xy}$ to the path-relative coordinate frame of  $\rho^*_{xy}$ to get the along-track and cross-track representations of the predicted trajectory $\hat{\tau}_{ac}=\Pi_{\rho^*}(\hat{\tau}_{xy})$ and the ground truth trajectory $\tau_{ac}=\Pi_{\rho^*}(\tau_{xy})$. The cross-track error of a prediction point $\hat{\tau}^{t}_{xy}$ is the absolute value of its cross-track component $\left|\hat{\tau}_{c}^t \right|$. The along-track error of the prediction point is the absolute difference between its along-track component and the along-track component of the corresponding ground truth point at the same timestamp $\left|\hat{\tau}_{a}^t - \tau_{a}^t \right|$. We use AATE and ACTE to refer to the average along-track error and average cross-track error, where the average is taken over all time horizons (analogous to ADE).


\section{Supplementary Results}
\label{app:results}

\subsection{Evaluation on Turning Cases}
\vspace{-.1cm}
In Table \ref{tab:turning-results}, we evaluate our method and baselines on a subset of our data which contains challenging turning cases. For an actor, if the heading deviation between the last future waypoint in the prediction horizon and the current heading is larger than 10 degrees, we define it as a turning behavior. For our internal dataset, these cases comprise 21.68\% of our test set; for the nuScenes validation set, this number is 24.28\%. Comparing with Table \ref{tab:results}, we notice that the performance gap between our method and other methods is even wider, suggesting that our model performs better especially on more challenging scenarios when actors are not simply driving straight. 


\begin{table}[h]
\begin{center}
\resizebox{\textwidth}{!}{
\begin{tabular}{lccccccc}
\toprule
Method & $\min_{1}$ADE  & $\min_{1}$FDE & $\min_{3}$ADE & $\min_{3}$FDE  & $\min_{5}$ADE  & $\min_{10}$ADE & $\mathbb{E}$[ADE] \\
\midrule
  MTP  & 3.67 (3.45) & 10.15 (9.71) & 2.70 (2.87) & 7.52 (8.10) & 2.70 (2.87) & 2.70 (2.87) & 3.81 (3.50) \\
  MultiPath  & 3.89 (4.60) & 10.70 (12.09) & 2.69 (3.01) & 7.53 (8.03) & 2.29 (2.47) & 1.87 (1.98) & 4.20 (4.70) \\
  CoverNet  & \, --- \, (5.53) & \, --- \, (13.07) & \, --- \, (3.87)  & \, --- \, (9.42) & \, --- \, (3.29) & \, --- \, (2.64) & \, --- \, (5.66) \\[1pt]
  \hdashline\\[-7pt]
  GoalNet-1T & \textbf{3.17} (\textbf{2.70}) & \textbf{8.59} (\textbf{7.39}) & 2.51 (2.19) & 6.56 (5.88) & 2.42 (2.16) & 2.42 (2.15) & \textbf{3.24} (\textbf{2.75}) \\
  GoalNet-2T  & 3.44 (3.04) & 9.30 (8.27) & \textbf{2.15} (\textbf{1.84}) & \textbf{5.67} (\textbf{4.74}) & \textbf{1.87} (\textbf{1.64}) & \textbf{1.76} (\textbf{1.58}) & 3.59 (3.14) \\
\bottomrule
\end{tabular}
}
\vspace{.2cm}
\caption{Trajectory prediction metrics on our internal dataset and on the public nuScenes dataset on {turning cases} only. Results listed as internal (nuScenes). All errors are reported in meters.}
\vspace{-.5cm}
\label{tab:turning-results}
\end{center}
\end{table}

\subsection{Evaluation on Best Matching Trajectory}
\vspace{-.1cm}

Finally, for completeness, we report results by comparing all methods on the trajectory error from the single trajectory that {best matches} the ground truth across \emph{all} trajectory modes. We call this the $\min_*$ metric. This comparison naturally provides an advantage to methods that generate more trajectories. Given that, we are not surprised to see in Table \ref{tab:best-matching-results} that CoverNet (which has 2206 modes) always achieves the best performance on nuScenes, and MultiPath (which has 64 modes) nearly always achieves the best performance on our internal dataset where CoverNet is not available. 

However, despite the fact that our method has many fewer modes than both MultiPath and CoverNet (see Table~\ref{tab:compactness}), we see that we are actually quite competitive on $\min_*$ACTE, which is the average cross-track error of the best-matching mode. In particular, we are able to outperform MultiPath on this metric (on both our dataset and nuScenes) and even come close to matching CoverNet's performance. Since the cross-track error captures the spatial component of the trajectory error, we attribute our good performance on this metric to our use of lane-based path anchors. Finally, we observe that this pattern is further emphasized when we evaluate only on turning cases.

\begin{figure}[h]
    \centering
    \begin{minipage}[t][][b]{.49\textwidth}
    \resizebox{\textwidth}{!}{
    \begin{tabular}{lccc}
    \toprule
    Method &  $\min_{*}$ADE  & $\min_{*}$AATE & $\min_{*}$ACTE  \\
    \midrule
    MTP  & 1.69 (2.10) & 1.44 (1.78) & 0.52 (0.71) \\
    MultiPath  & \textbf{0.95} (0.97) & \textbf{0.75} (0.74) & {0.41} (0.46) \\
    CoverNet  & \, --- \, (\textbf{0.75}) & \, --- \, (\textbf{0.58}) & \, --- \, (\textbf{0.35})  \\[1pt]
    \hdashline\\[-7pt]
    GoalNet-1T & 1.79 (1.75) & 1.61 (1.57) & 0.42 (0.47) \\
    GoalNet-2T  & 1.28 (1.22) & 1.12 (1.05) & \textbf{0.36} (0.42) \\
    \bottomrule
    \end{tabular}
    }
    \end{minipage}
    \hfill
    \begin{minipage}[t][][b]{.49\textwidth}
    \resizebox{\textwidth}{!}{
    \begin{tabular}{lccc}
    \toprule
    Method &  $\min_{*}$ADE  & $\min_{*}$AATE & $\min_{*}$ACTE  \\
    \midrule
    MTP  & 2.70 (2.87) & 2.05 (2.22) & 1.18 (1.28) \\
    MultiPath  & \textbf{1.51} (1.45) & \textbf{1.04} (0.98) & 0.86 (0.86) \\
    CoverNet  & \, --- \, (\textbf{1.00}) & \, --- \, (\textbf{0.70}) & \, --- \, (\textbf{0.57})  \\[1pt]
    \hdashline\\[-7pt]
    GoalNet-1T & 2.42 (2.15) & 2.07 (1.79) & 0.73 (0.77) \\
    GoalNet-2T  & 1.77 (1.58) & 1.44 (1.26) & \textbf{0.65} (0.68) \\
    \bottomrule
    \end{tabular}
    }
    \end{minipage}
    \captionof{table}{Trajectory prediction metrics of the best overall trajectory ($\min_*$) produced by each method on both our internal dataset and on the public nuScenes dataset. The left table shows results on all samples, and the right table shows results on turning cases. Results listed as internal (nuScenes). All errors are reported in meters.}
   \label{tab:best-matching-results}
\end{figure}

\end{appendices}

\end{document}